\documentclass[sn-mathphys,Numbered]{sn-jnl}
\usepackage{xcolor}
\usepackage{booktabs}
\usepackage{graphicx}
\usepackage{multirow}
\usepackage{manyfoot}
\usepackage{amsmath,amssymb,amsfonts}

\usepackage[ruled,linesnumbered]{algorithm2e}

\begin{document}
\title[Article Title]{Leveraging Deep Neural Networks for Aspect-Based Sentiment Classification}
%==========
\author[1]{\fnm{Chen} \sur{Li}}
\author[2]{\fnm{Debo} \sur{Cheng}}
\author[3]{\fnm{Yasuhiko} \sur{Morimoto}}
%==========
\affil[1]{\orgdiv{D3 Center}, \orgname{Osaka University}, \country{Japan}}
\affil[2]{\orgdiv{UniSA STEM}, \orgname{University of South Australia}, \country{Australia}}
\affil[3]{\orgdiv{Graduate School of Advanced Science and Engineering}, \orgname{Hiroshima University}, \country{Japan}}

%****************************************************************
\abstract{
Aspect-based sentiment analysis seeks to determine sentiment with a high level of detail. While graph convolutional networks (GCNs) are commonly used for extracting sentiment features, their straightforward use in syntactic feature extraction can lead to a loss of crucial information. This paper presents a novel \underline{\bf e}dge-\underline{\bf e}nhanced \underline{\bf GCN}, called EEGCN, which improves performance by preserving feature integrity as it processes syntactic graphs. We incorporate a bidirectional long short-term memory (Bi-LSTM) network alongside a self-attention-based transformer for effective text encoding, ensuring the retention of long-range dependencies. A bidirectional GCN (Bi-GCN) with message passing then captures the relationships between entities, while an aspect-specific masking technique removes extraneous information. Extensive evaluations and ablation studies on four benchmark datasets show that EEGCN significantly enhances aspect-based sentiment analysis, overcoming issues with syntactic feature extraction and advancing the field's methodologies.}

\maketitle

%****************************************************************
\section{Introduction}
\label{sec:intro}
Aspect-based sentiment analysis focuses on determining sentiment at a more granular level, where different aspects within the same sentence can exhibit distinct sentiments. For example, consider the restaurant review: “Although the menu is limited, the friendly staff provided us with a nice night." The sentiment associated with the “menu" aspect is negative (“limited"), while the "staff" aspect conveys a positive sentiment (“friendly"). This situation, where a single sentence contains multiple aspects and sentiments, complicates sentiment prediction. Addressing these complexities is a growing area of research, with ongoing efforts on enhancing analytical methods \cite{bauman2022know}.

Neural networks have proven effective in aspect-based sentiment analysis due to their ability to model sequences, with recurrent neural networks (RNNs) \cite{zhang2019multi} and Transformers \cite{vaswani2017attention} being prominent examples \cite{wang2016attention,tang2015effective,chen2017recurrent,ma2017interactive,fan2018multi}. Nonetheless, these models often face challenges in accurately capturing the relationships between aspects and sentiment-bearing words. For instance, in the sentence “The dish looks mediocre but tastes surprisingly wonderful," although both sentiments “mediocre" and “tastes surprisingly wonderful" are related to the aspect “dish," the latter is more relevant. This nuanced understanding is challenging for these models due to proximity issues. Convolutional neural networks (CNNs) \cite{rani2022efficient} have been explored to better capture sentiment in phrases. However, CNNs are limited by their local receptive fields, making it difficult to handle long-range dependencies, such as distinguishing the sentiment expressed in “tastes surprisingly wonderful" from just “wonderful" \cite{kipf2016semi}.

\begin{figure}[t]
\centering  
\includegraphics[width=0.8\hsize]{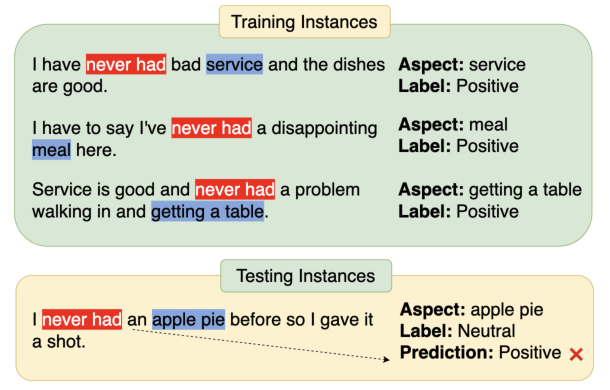}   
\caption{An example of spurious associations of a customer review system.}
\label{fig:example} 
\end{figure}

Graph convolutional networks (GCNs) \cite{poria2014dependency,zhou2020graph} can represent sentences using a tree-like structure of entities, creating more direct connections between aspects and associated words. This approach helps address long-distance dependencies and aids in learning node representations and their positional relationships \cite{meng2020structure}. Nevertheless, standard GCNs often lack edge information, which limits their ability to capture detailed syntactic and semantic relationships, leading to suboptimal performance \cite{chang2023reducing,scaria2023instructabsa}. For example, as illustrated in Figure \ref{fig:example}, while the aspect “service" might be linked to the phrase “never had" with positive sentiment in past reviews, a new review may present a different sentiment. The network's reliance on "never had" might fail to accurately reflect the sentiment in this new context.

This study builds upon our prior research \cite{li2024adma} by introducing EEGCN, a novel \underline{\bf e}dge-\underline{\bf e}nhanced (\underline{\bf GCN}) designed to advance aspect-based sentiment analysis. Our approach incorporates bidirectional long short-term memory (Bi-LSTM) for node word embedding extraction and utilizes Bi-GCN to model dependencies within a tree structure, thereby refining word representations. This integration improves the accurate mapping of sentiment words to target aspects, leading to significant performance gains. Extensive evaluation experiments validate the robustness of EEGCN, showcasing its superior capabilities. The main contributions are as follows:
\begin{itemize} 
\item {\bf Neural Network Architecture:} EEGCN integrates Bi-LSTM, a transformer, and an edge-enhanced Bi-GCN. This combination utilizes linguistic syntax to form a graph that supports advanced textual representation and sentiment decoding, marking a significant step forward in sentiment analysis techniques.
\item {\bf Enhanced Semantic Edges:} By using a dependency tree, EEGCN minimizes the distance between aspects and their related words, establishing a clearer syntactic path. The model incorporates edge information from these trees to strengthen the syntactic ties between words and their target aspects.
\item {\bf Improved Performance:} The effectiveness of the proposed method is demonstrated through testing on four standard benchmark datasets, where EEGCN consistently outperforms other baseline models in most performance metrics. 
\end{itemize}

This paper follows a structured organization outlined as follows: We present a survey of related studies in Section \ref{sec:related}. Section \ref{sec:model} provides a detailed exposition of our EEGCN model, elucidating its components and functionalities. In Section \ref{sec:exp}, we perform comprehensive experiments to validate the efficacy of EEGCN. Furthermore, we conduct ablation studies to meticulously explore the individual effects of each model component. Finally, we conclude the paper in Section \ref{sec:conclusion}.

%****************************************************************
\section{Related Work}
\label{sec:related}
In this section, we first conduct a thorough examination of convolutional aspect-based sentiment analysis systems. Subsequently, acknowledging their inherent limitations, we shift our focus to neural network-based models, with a particular emphasis on the rising prominence of sentiment classification models rooted in RNNs over recent years. Finally, we conclude this section with an exploration of advanced GCN-based neural architectures specifically designed for sentiment analysis systems.

%-----------------------------------
\subsection{Aspect-based Sentiment Classification}
Traditional sentiment analysis often relies on classifiers trained with features like bag-of-words and sentiment dictionaries \cite{rao2009semi}. These methods, including both rule-based \cite{ding2008holistic} and statistical approaches \cite{jiang2011target}, are constrained by the need for handcrafted features, which limits their effectiveness in analyzing the complex nature of customer reviews. Specifically, these traditional methods struggle to capture the nuanced sentiments associated with individual aspects of interest.

In response to the limitations of conventional approaches, there has been a significant shift toward aspect-based sentiment analysis. This field focuses on four primary tasks \cite{cai2020aspect}: aspect term sentiment analysis (ATSA) \cite{manek2017aspect}, aspect category sentiment analysis (ACSA) \cite{zhu2019aspect}, aspect term extraction (ATE) \cite{ma2019exploring}, and aspect category extraction (ACE) \cite{kumar2021sharing}. ATSA aims to determine the sentiment polarity associated with specific aspect terms, providing a detailed understanding of sentiment in relation to particular review elements \cite{phan2020modelling}. ACSA focuses on predicting sentiment polarity within predefined categories, offering insights into broader sentiment patterns tied to these categories \cite{liao2021improved}. ATE and ACE address the extraction of aspects from reviews, with ATE concentrating on precise aspect terms, such as “dish,” and ACE dealing with broader categories, like “food,” which includes various terms like “dish” \cite{augustyniak2021comprehensive}.

The goal of these tasks is to achieve a more precise and contextually relevant understanding of sentiment expressions. By enhancing the analysis of sentiments linked to specific aspects, aspect-based sentiment analysis improves accuracy and relevance in sentiment analysis applications \cite{do2019deep}. Nevertheless, challenges such as handling ambiguous language and context remain, highlighting the need for ongoing research to advance aspect-based sentiment analysis methodologies.

%-----------------------------------
\subsection{RNN-based Sentiment Classification}
Recent advancements in neural networks have spurred significant interest in various applications \cite{li2018capturing,li22transformer,li2023spotgan,li2024gxvaes}. For target-dependent sentiment classification, models like those proposed in \cite{tang2016effective} leverage neural networks, with LSTM architectures proving effective for aspect-based sentiment analysis. These models, known as target-dependent LSTM (TD-LSTM), integrate target-specific information to enhance the understanding of how a target word interacts with its context, capturing relevant context segments to determine sentiment polarity. TD-LSTM can be trained end-to-end using backpropagation with cross-entropy loss, but it may struggle with capturing subtle sentiment nuances due to its focus on target-specific interactions.

Building on the TD-LSTM framework, ATAE-LSTM \cite{wang2016attention} incorporates attention mechanisms with Bi-LSTM to emphasize contextual relationships. While these models improve contextual understanding, they may still face challenges in capturing long-range dependencies when crucial details are far from the target. To address this, RAM \cite{chen2017recurrent} and TNet-LF \cite{li2018transformation} introduce multi-attention and Bi-attention mechanisms, respectively, which enhance their capacity to handle extensive contexts. However, TNet-LF's limited ability to capture detailed sentence structure and grammar can affect its performance, especially in complex syntactic scenarios. Sentic LSTM \cite{ma2018targeted} integrates commonsense knowledge into an attentive LSTM network for targeted aspect-based sentiment analysis, while RACL \cite{chen2020relation} employs relation-aware collaborative learning for comprehensive sentiment analysis tasks. Despite its broad scope, RACL's effectiveness remains unclear due to a lack of detailed evaluations compared to existing models.

The Transformer \cite{vaswani2017attention,li2024tengan} represents a significant advancement, designed to excel in capturing complex long-range dependencies in text. Its proven effectiveness makes it a strong candidate for tasks requiring thorough contextual analysis. 

%-----------------------------------
\subsection{GCN-based Sentiment Classification}
Unlike RNN-based sentiment analysis systems, GCN-based approaches represent language data as graphs, where nodes correspond to words or phrases and edges denote their relationships. This graph-based representation provides a more detailed understanding of contextual dependencies, capturing complex connections that traditional RNNs may overlook \cite{wang2020relational}.

Recent developments in GCN-based aspect-based sentiment analysis include several notable models. ASGCN \cite{zhang2019aspect} excels at capturing sentiment polarity across different aspects by modeling relationships between aspects and their related words with GCNs. However, ASGCN's performance can be hindered by inaccuracies in dependency tree parsing, which may compromise the integrity of aspect-word relationships and overall effectiveness. AEGCN \cite{xiao2020targeted} combines GCNs with attention mechanisms, such as multi-head interactive and self-attention, to enhance sentiment analysis. Yet, any errors in attention encoding can negatively impact the accuracy of the graph's relationships. SK-GCN \cite{zhou2020sk} integrates a syntactic dependency tree with GCNs to improve syntax modeling. Nonetheless, it struggles with limitations related to the representation of commonsense knowledge due to insufficient graph structures, affecting its performance in aspect-level sentiment analysis. AGCN \cite{zhao2022aggregated} aggregates features from neighboring nodes to enhance node representations. However, its effectiveness is highly dependent on the accuracy of the graph structure; if the graph fails to capture relationships, AGCN may produce less meaningful features, leading to diminished performance.

Unlike previous approaches, we introduce a novel integration of Bi-LSTM, Transformer, and GCNs to enhance aspect-based sentiment analysis. EEGCN leverages the unique strengths of each component: Bi-LSTM effectively extracts features by capturing detailed patterns and semantic nuances within the text. The Transformer excels in managing long-range dependencies, which is essential for understanding the broader contextual relationships crucial for accurate sentiment analysis. Additionally, by incorporating a GCN with a dependency parsing layer, EEGCN integrates syntactic information, providing a more detailed understanding of syntactic relationships within the text. This combined approach aims to deliver a more comprehensive and nuanced perspective on aspect-based sentiment analysis, addressing the limitations of previous methods and advancing the overall accuracy of sentiment analysis methodologies.

%***************************************************************
\section{Model}
\label{sec:model}

\subsection{Architecture Overview}
\begin{figure}[t]  
\centering  
\includegraphics[width=0.94\hsize]{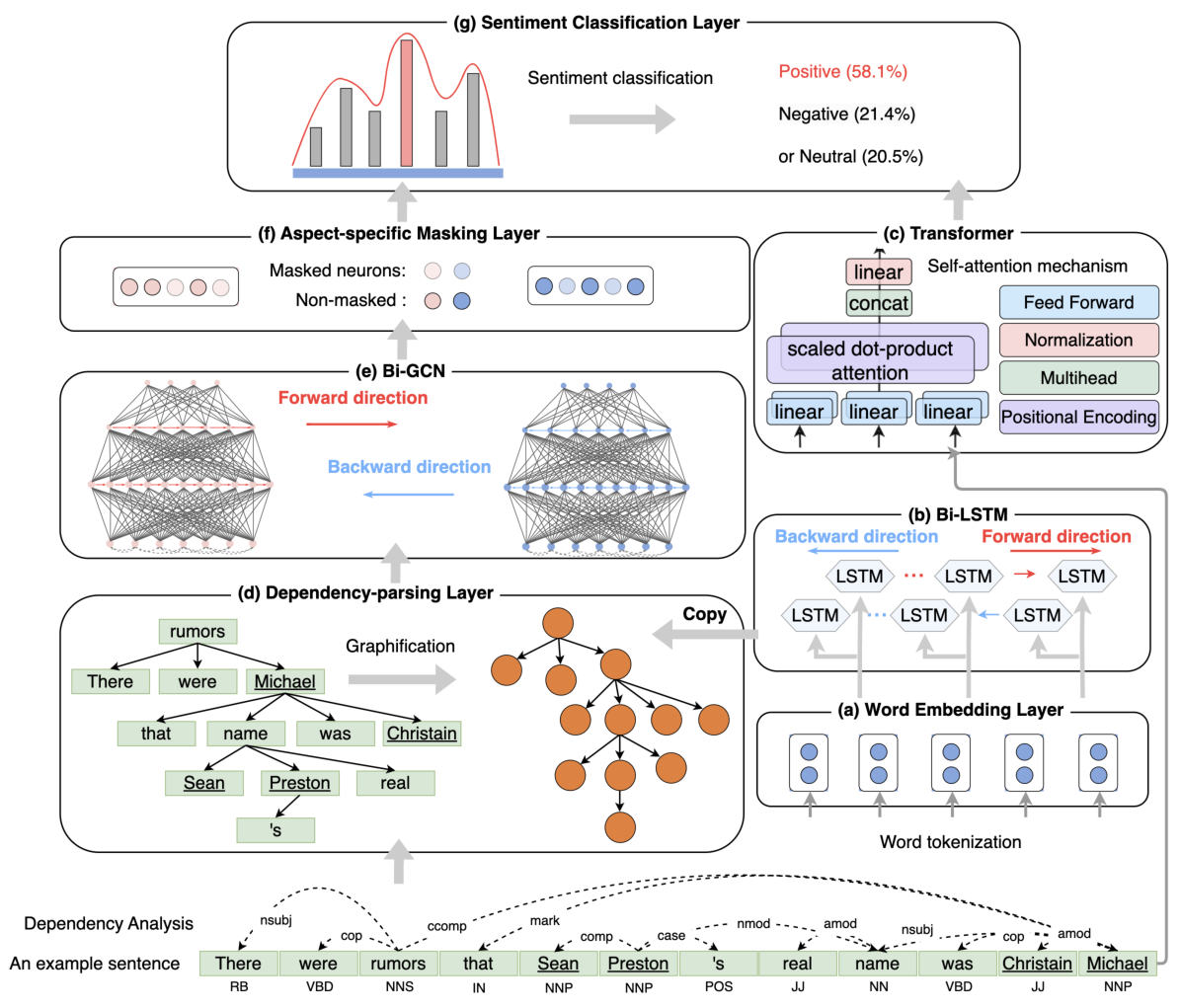}   
\caption{The EEGCN architecture for aspect-based sentiment analysis consists of the following key components: (a) The word embedding layer converts sentences into vector representations, forming the basis for further processing. (b) The Bi-LSTM network captures contextual relationships within the sentence to provide a deeper understanding of word interactions. (c) The transformer layer, with its self-attention mechanism, analyzes global word features and interrelationships in lengthy texts, enhancing contextual comprehension. (d) The dependency parsing layer constructs a dependency tree to represent grammatical structures and word dependencies. (e) The Bi-GCN utilizes message passing to propagate information across the nodes of the dependency tree, effectively modeling word relationships. (f) The aspect-specific masking technique refines representations by isolating aspect-related information, reducing redundancy and improving analysis accuracy. (g) The classification layer categorizes the sentiment of the sentence into predefined categories, such as positive, negative, or neutral.}
\label{fig:overview} 
\end{figure}

In this section, we describe the proposed EEGCN in seven key components. First, the word embedding layer generates low-dimensional representations for each token in the vocabulary. The Bi-LSTM then extracts sentence features by capturing intricate patterns. The Transformer model contributes by analyzing contextual features from a global perspective. A dependency parsing layer constructs a tree structure to represent grammatical relationships. The Bi-GCN processes this tree structure to integrate syntactic information. An aspect-specific masking layer minimizes redundancy and improves accuracy. Finally, the sentiment classification layer determines the sentiment of the sentence. Figure \ref{fig:overview} illustrates the overall architecture of EEGCN for aspect-based sentiment analysis.

%-----------------------------------
\subsection{EEGCN}
To tokenize a sentence comprising $n$ words with an $m$-word aspect, we represent it as a sequence denoted by ${\bf S}$ in the following manner:
\begin{align}
\label{eq:tokenize}
{\bf S} = \left\{{\bf w}_1, {\bf w}_2, \cdot\cdot\cdot, {\bf w}_{\gamma}, \cdot\cdot\cdot, {\bf w}_{\gamma+m}, \cdot\cdot\cdot, {\bf w}_n \right\},
\end{align}
where ${\bf w}_t \in {\bf S}$ represents a word token and $\gamma$ denotes the starting index of the aspect. Each word token is subsequently mapped to a low-dimensional vector ${\bf e}_t \in \mathbb{R}^{d_w}$, with $d_w$ indicating the dimensionality of the word embedding.

%-----------------------------------
Word embedding maps the words in the sentence to low-dimensional vectors, which are then fed into the Bi-LSTM network \cite{graves2013speech}. The Bi-LSTM network facilitates the extraction of contextual information from sentences. As outlined in \cite{hu2021graph}, the Bi-LSTM consists of two parts: the forward LSTM and the backward LSTM. In the forward LSTM, the hidden vector $\overrightarrow {\bf h}^{LSTM}_t$ at the current time step is computed by combining the input embedding ${\bf e}t$ with the hidden vector $\overrightarrow {\bf h}^{LSTM}{t-1}$ from the previous time step. The backward LSTM processes ${\bf e}t$ and incorporates the hidden vector $\overleftarrow {\bf h}^{LSTM}{t+1}$ from the following time step. Ultimately, the hidden vectors $\overrightarrow {\bf h}^{LSTM}_t$ and $\overleftarrow {\bf h}^{LSTM}_t$ are concatenated to form the combined hidden vector ${\bf h}^{LSTM}_t$. This procedure is depicted as follows:
\begin{align}
\label{eq:lstm1}
\overrightarrow {\bf h}^{LSTM}_t = \text{LSTM}({\bf e}_t, \overrightarrow {\bf h}^{LSTM}_{t-1}),~\overrightarrow {\bf h}^{LSTM}_t \in \mathbb{R}^{d_h},\\
\setlength{\abovedisplayskip}{0pt}
\label{eq:lstm2}
\overleftarrow {\bf h}^{LSTM}_t = \text{LSTM}({\bf e}_t, \overleftarrow {\bf h}^{LSTM}_{t+1}),~\overrightarrow {\bf h}^{LSTM}_t \in \mathbb{R}^{d_h},\\
\setlength{\abovedisplayskip}{0pt}
\label{eq:lstm3}
{\bf h}^{LSTM}_t = \overrightarrow {\bf h}^{LSTM}_t \oplus \overleftarrow {\bf h}^{LSTM}_t,~{\bf h}^{LSTM}_t \in \mathbb{R}^{2 \times d_h},
\end{align}
where $d_h$ denotes the dimension of the hidden layer in the Bi-LSTM, and $\oplus$ represents the concatenation operator.

%-----------------------------------
A transformer encoder, as detailed in \cite{vaswani2017attention}, employs self-attention mechanisms to analyze relationships and global features of words within long sentences. To account for the sequential positions within a sentence, sinusoidal positional encoding functions are used, as described below:
\begin{align}
\label{eq:trans1}
\text{P}_{pos, 2i} = \sin (\frac{pos}{10000^{\frac{2i}{d_{model}}}})~~\text{and}~~\text{P}_{pos, 2i+1} = \cos (\frac{pos}{10000^{\frac{2i}{d_{model}}}}),
\end{align}
where $pos$ indicates the position of a word in the sentence, and $i$ refers to the $i$th dimension of the embedding, with the embedding dimension specified as $d_{model}$. The input to the transformer is the vector formed by combining the word embedding with positional encoding. The transformer then computes attention weights as follows:
\begin{align}
\label{eq:trans3}
\text{Attention}({\bf Q, K, V}) = \text{Softmax}(\frac{{\bf QK}^T}{\sqrt{d_k}}){\bf V},
\end{align}
where ${\bf Q}$, ${\bf K}$, and ${\bf V}$ denote the query, key, and value matrices with dimensions $d_k$, $d_k$, and $d_v$, respectively. Typically, a transformer utilizes a multi-head attention mechanism to project the query, key, and value matrices $h$ times, enhancing internal contextual relationships. The mathematical formulation for this process is as follows:
\begin{align}
\label{eq:trans4}
&\text{Multi-Head}({\bf Q, K, V}) = [\text{Head}_1, \cdots, \text{Head}_h] {\bf W},\\
\setlength{\abovedisplayskip}{0pt}
\label{eq:trans5}
&\text{Head}_i = \text{Attention}({\bf QW}_Q, {\bf KW}_K, {\bf VW}_V),
\end{align}
where ${\bf W}$, ${\bf W}_Q$, ${\bf W}_K$, and ${\bf W}V$ represent the weight matrices used in the self-attention mechanism. Subsequently, the transformer encoder applies two normalization layers to extract deep contextual features. Let ${\bf Z}{out}$ denote the output matrix produced by the transformer encoder, which can be formulated as follows:
\begin{align}
\label{eq:trans6}
{\bf Z}_{out} = \text{Transformer}({\bf S}).
\end{align}
Finally, the global attention scores computed are crucial for the final sentiment classification layer. The output hidden vector from the Bi-LSTM network, denoted as ${\bf h}_t$, serves as an input to the dependency-parsing layer within the Bi-GCN, which helps identify word dependencies by constructing a dependency tree. Formally, the Bi-GCN starts by initializing a matrix to record edge weights. It first generates the adjacency matrix ${\bf A} \in \mathbb{R}^{n \times n}$ from the dependency tree of the sentence. The standard form of this adjacency matrix is given by:
\begin{equation}
\label{eq:parse1}
{\bf A}_{ij} =
\begin{cases}
1, \quad i=j. \\
1, \quad i \ne j,~i~\text{and}~j~\text{have dependencies}. \\
0, \quad \text{otherwise}.
\end{cases}
\end{equation}
GCNs struggle to capture detailed edge information due to the discrete nature of the adjacency matrix (which contains only 0s and 1s). To address this, we propose an adjacency matrix that includes non-discrete edge information, formulated as follows:
\begin{equation}
\label{eq:parse2}
{\bf A} _{ij} =
\begin{cases}
1, \quad i=j, \\
\text{SDI}(i,j), \quad i \ne j,~i~\text{and}~j~\text{have dependencies}, \\
0, \quad \text{otherwise}.
\end{cases}
\end{equation}
Here, SDI is employed to calculate the syntactic dependency information between word $i$ and word $j$. The computation is summarized as follows:
\begin{align}
\label{eq:parse3}
\text{SDI}(i, j)= \frac{\text{Count}(\text{sd}(i, j))}{\text{Count}(\text{sd}(\cdot))}, 
\end{align}
where $\text{sd}(i, j)$ denotes the syntactic relationship between word pair $i$ and $j$, and $\text{sd}(\cdot)$ represents the overall syntactic information of the dataset.

%-----------------------------------
A GCN is a specialized variant of CNNs tailored for handling and encoding structured graph data. In the case of a textual graph with $n$ words, we generate an adjacency matrix from the syntactic matrix $\mathbf{A} \in \mathbb{R}^{n \times n}$ to encode syntactic dependencies. The representation of word $i$ at layer $l$ is denoted by ${\bf h}^l_i$. This graphical representation is conceptually depicted as follows:
\begin{equation}
\label{eq:gcn1}
{\bf h}^l_i = \sigma (\sum_{j=1}^{n}\tilde{\mathbf{A}}_{ij}\mathbf{W}^{l}h_{j}^{l-1}+\mathbf{b}^{l}).
\end{equation}
In the GCN, $\mathbf{W}^l$ represents the weight matrix for linear transformations, $\mathbf{b}^l$ is the bias term, and $\sigma$ is the nonlinear activation function. By utilizing the sentence’s dependency tree, the GCN applies syntactic constraints to help identify words related to a specific aspect, even when they are not adjacent. This capability makes the GCN particularly effective for aspect-based sentiment analysis. The update of each word's representation is performed through the Bi-GCN using the following process:
\begin{equation}
\label{eq:gcn2}
\overrightarrow {\bf h}^l_i =\sum_{j=1}^{n}\tilde{\mathbf{A}}_{ij}\mathbf{W}^{l}h_{j}^{l-1}~~\text{and}~~
\overleftarrow {\bf h}^l_i = \sum_{j=1}^{n}{\tilde{\mathbf{A}}_{ij}}^{T} \mathbf{W}^{l}h_{j}^{l-1},
\end{equation}
where $\overrightarrow{\bf h}^l_i$ and $\overleftarrow{\bf h}^l_i$ represent the outputs of the forward and backward hidden layers for the word $i$ at layer $l$. These forward and backward representations can be combined by concatenating them as follows:
\begin{equation}
\label{eq:gcn4}
\tilde{\bf h}^l_i = \overrightarrow {\bf h}^l_i \oplus \overleftarrow {\bf h}^l_i
~~\text{and}~~
{\bf h}^l_i = \text{ReLU}(\frac{\tilde{\bf h}^l_i}{d_i+1}\mathbf{W}^{l}+\mathbf{b}^{l}),
\end{equation}
where $d_i$ denotes the degree of the $i$th token in the adjacency matrix. It is important to note that the parameters $\mathbf{W}^{l}$ and $\mathbf{b}^{l}$ are trainable within the Bi-GCN network.

%-----------------------------------
We apply a masking technique to minimize redundancy in the hidden representations generated by the Bi-GCN by filtering out aspect-independent text. Specifically, this masking process can be described as follows:
\begin{equation}
\label{eq:mask1}
{\bf h}_i^l = 0,~1\leq i< \gamma,~\gamma +m-1<t\leq n.
\end{equation}
The result from the zero mask layer consists of aspect-focused characteristics.
\begin{equation}
\label{eq:mask2}
\mathbf{H}_{mask}^l = \left\{{\bf 0}, \cdots, {\bf h}_{\gamma}^l , \cdots, {\bf h}_{\gamma+m-1}^l, \cdots, {\bf 0} \right\}.
\end{equation}

Through graph convolution, the feature vector $\mathbf{H}_{mask}^l$ effectively captures the contextual information related to specific aspects, integrating both syntactic dependencies and extended multi-word relationships. Additionally, an attention mechanism is utilized to generate a vector representation of the masked layer. The calculation of attention weights is carried out as follows:
\begin{equation}
\label{eq:mask3}
\beta_i=\sum_{j=1}^{n}{\bf h}_j^T {\bf h}_j^l~~\text{and}~~
\alpha_i=\frac{\exp(\beta_i)}{\sum_{j=1}^n\exp(\beta_j))}.
\end{equation}
Finally, the sentiment classification combines the aggregated representation scores with the attention scores produced by the transformer. This results in the following sentiment classification:
\begin{equation}
\label{eq:mask5}
\mathbf{res}_{out} = \sum_{i=1}^n\alpha_i{\bf h}_i + {\bf Z}_{out}.
\end{equation}

%-----------------------------------
The $\mathbf{r}_{out}$ obtained from the previous step is fed into the fully connected layer for sentiment classification. The formula is as follows:
\begin{equation}
\label{eq:classifer}
\mathbf{prob}= \text{softmax}(\mathbf{res}_{out}\mathbf{W}_p + \mathbf{b}).
\end{equation}

\subsection{Training Phase}
\label{sec:training}
Using both cross-entropy and $\mathit{L_{2}}$ regularization \cite{cortes20092}, the loss function is as follows:
\begin{equation}
\label{eq:train}
\text{Loss} = - \sum_{\hat{\mathit{p}}\in C}\log\mathbf{prob_{\hat{\mathit{p}}}}+ \lambda \left \| \Theta  \right \|_{2},
\end{equation}
where $\mathit{p}$ denotes the label and $C$ represents the dataset. Here, $\mathbf{prob_{\hat{\mathit{p}}}}$ is the $\hat{\mathit{p}}$th element of $\mathbf{prob}$. The parameters $\Theta$ are the trainable parameters, and $\lambda$ is the coefficient for $\mathit{L_{2}}$ regularization. For a clearer understanding of the EEGCN model, Algorithm \ref{alg:algorithm} details the training procedure, offering a more thorough insight into its operation.

% Algorithm
\begin{algorithm}[t]
\caption{Algorithm for the proposed EEGCN.}
\label{alg:algorithm}
\tcp{Read sample sentences from datasets.}
Tokenize the sentence using Eq.(\ref{eq:tokenize}).\\
Embed the words using the word embedding layer.\\
Examine the dependency information within the sentence.\\
\tcp{A Bi-LSTM layer.}
Apply a Bi-LSTM layer to capture the contextual features of the sample sentence using Eqs.(\ref{eq:lstm1}),(\ref{eq:lstm2}), and (\ref{eq:lstm3}).\\
\tcp{A transformer layer.}
Encode positional information using Eq.(\ref{eq:trans1}).\\
Utilize a transformer encoder architecture to capture the global features within a sentence, following Eqs.(\ref{eq:trans3}), (\ref{eq:trans4}), (\ref{eq:trans5}), and (\ref{eq:trans6}).\\
\tcp{Dependency-parsing layer.}
Replace the weight matrix in Eq.(\ref{eq:parse1}) with Eq. (\ref{eq:parse2}) to retain additional edge information for the graph.\\ 
Compute SDI using Eq.(\ref{eq:parse3}).
\tcp{A Bi-GCN submodule.}
Employ Bi-GCN to extract features from the structural graph using Eqs.(\ref{eq:gcn1}), (\ref{eq:gcn2}), and (\ref{eq:gcn4}).\\
\tcp{Aspect-specific masking.}
Apply aspect-specific masking techniques to minimize redundant information in the Bi-GCN, as outlined in Eqs.(\ref{eq:mask1}), (\ref{eq:mask2}), (\ref{eq:mask3}), and (\ref{eq:mask5}).\\
\tcp{Sentiment classification.}
Use fully connected layers to build a classifier for sentiment categorization as specified in Eq.(\ref{eq:classifer}).\\
\tcp{Training loss.}
Train EEGCN using the formula provided in Eq.(\ref{eq:train}).\\
\tcp{Testing phase.}
Assess the effectiveness of EEGCN using test datasets.
\end{algorithm}

%****************************************************************
\section{Experiments}
\label{sec:exp}
In this section, we validate the effectiveness of the EEGCN model through experiments conducted on four commonly used benchmark datasets. Additionally, the impact of GCN layers on performance is analyzed. Finally, ablation studies are performed.

%-----------------------------------
\subsection{Datasets}
\label{sec:datasets}
The EEGCN model's performance was evaluated using four widely recognized benchmark datasets. The Twitter dataset \cite{dong2014adaptive} includes 6,940 posts from Twitter users, split into 6,248 for training and 692 for testing. Additionally, three datasets—Rest14 \cite{pontiki-etal-2014-semeval}, Rest15 \cite{pontiki2015semeval}, and Rest16 \cite{pontiki2016semeval}—feature customer comments and restaurant ratings. Rest14 consists of 4,728 comments, with 3,608 for training and 1,120 for testing. Rest15 has 1,746 reviews, divided into 1,204 for training and 542 for testing. Rest16 contains 2,364 reviews, with 1,748 used for training and 616 for testing. To improve dataset quality, sentences with ambiguous or conflicting sentiments were excluded from the Rest15 and Rest16 datasets. A detailed summary is provided in Table \ref{tab:statistics}.

\begin{table*}[t]
\caption{Statistics for the four datasets used to evaluate EEGCN.}
\label{tab:statistics}
\centering
\begin{tabular}{ccccc}
\toprule
Dataset & Split & Positive & Neutral & Negative \\\midrule
\multirow{2}[4]{*}{Twitter} & Train & 1561  & 3127  & 1560 \\   
& Test  & 173   & 346   & 173 \\\midrule
\multirow{2}[4]{*}{Rest14} 
& Train & 2164 & 637 & 807 \\         
& Test  & 728   & 196   & 196 \\\midrule
\multirow{2}[4]{*}{Rest15} & Train & 912   & 36    & 256 \\
& Test  & 326   & 34    & 182 \\\midrule
\multirow{2}[4]{*}{Rest16} & Train & 1240  & 69    & 439 \\
& Test  & 469   & 30    & 117 \\\bottomrule
\end{tabular}
\end{table*}

%-----------------------------------
\subsection{Experimental Setup}
\label{sec:setting}
For the experiments, we set up the word embedding layer with pre-trained GloVe vectors \cite{pennington2014glove}, featuring a dimension of 300. Similarly, the Bi-LSTM hidden layer is configured with a dimension of 300. The Bi-GCN within EEGCN comprises 3 layers, with parameters initialized from a uniform distribution. We use the Adam optimizer \cite{zhang2018improved} with a learning rate of 0.001, apply $\mathit{L_{2}}$ regularization, and process data in batches of 32. The model is trained for up to 100 epochs.

\begin{table}[t]
\caption{Result comparison of baseline models with EEGCN. Values highlighted in gray represent the highest scores.}
\label{tab:result}
\centering
\renewcommand\arraystretch{0.4}
\renewcommand\tabcolsep{1.5pt} 
%\resizebox{\textwidth}{16mm}{
\begin{tabular}{@{}lllllllll@{}}\toprule
\multirow{2}{*}{Model} & \multicolumn{2}{l}{Twitter} & \multicolumn{2}{l}{Rest14} & \multicolumn{2}{l}{Rest15} & \multicolumn{2}{l}{Rest16} \\ \cmidrule(l){2-9} 
&Acc (\%) $\uparrow$ &F1 (\%) $\uparrow$&Acc (\%) $\uparrow$&F1 (\%) $\uparrow$&Acc (\%) $\uparrow$&F1 (\%) $\uparrow$&Acc (\%) $\uparrow$&F1 (\%) $\uparrow$\\ \cmidrule(r){1-9}
SVM&63.40&63.30&80.16&-&-&-&-&-\\
LSTM&69.56&67.70&78.13&67.47&77.37&55.17&86.80&63.88\\
TD-LSTM&70.80&69.00&78.00&66.73&76.39&58.70&82.16&54.21\\
MemNet&71.48&69.90&79.61&69.64&77.31&58.28&85.44&65.99\\
AOA&72.30&70.20&79.97&70.42&78.17&57.02&87.50&66.21\\
IAN&72.50&70.81&79.26&70.09&78.54&52.65&84.74&55.21\\
TNet-LF&72.98&71.43&80.42&71.03&78.47&59.47&\colorbox[gray]{0.9}{89.07}&\colorbox[gray]{0.9}{70.43}\\
ASGCN&71.53&69.68&80.86&72.19&79.34&60.78&88.69&66.64\\ 
MGAN&72.54&70.81&81.25&71.94&79.36&57.26&87.06&62.29\\
Sentic LSTM&70.66&67.87&79.43&70.32&79.55&60.56&83.01&68.22\\
CMLA-ALSTM&-&-&77.46&63.87&81.03&54.79&-&-\\
RACL&-&-&81.42&69.59&\colorbox[gray]{0.9}{83.26}&59.85&-&-\\
AEGCN&73.16&71.82&81.04&71.32&79.95&60.87&87.39&68.22\\
SK-GCN&71.97&70.22&80.36&70.43&80.12&60.70&85.17&68.08\\
AGCN&\colorbox[gray]{0.9}{73.98}&\colorbox[gray]{0.9}{72.48}&80.02&71.02&80.07&\colorbox[gray]{0.9}{62.70}&87.98&65.78\\\midrule
EEGCN &72.44&70.67&\colorbox[gray]{0.9}{81.70}&\colorbox[gray]{0.9}{73.63}&79.34&61.99&88.80&68.96\\\toprule   
\end{tabular}%}
\end{table}

%-----------------------------------
\subsection{Metrics}
\label{sec:measures}
We evaluate the effectiveness of EEGCN using two key performance metrics: Accuracy (Acc) and Macro F1 (F1). Accuracy measures the proportion of correctly classified samples, reflecting the model's overall precision. Macro F1 accounts for both precision and recall, providing a balanced view of the model's performance.

%-----------------------------------
\subsection{Results}
\label{sec:results}
Table \ref{tab:result} provides a comparative analysis of Accuracy (Acc) and Macro F1 (F1) scores between the proposed EEGCN model and 15 other baseline models. The results indicate that EEGCN consistently outperforms its competitors across the Twitter, Rest14, and Rest15 datasets. Although the AGCN model achieved the highest Acc and F1 scores in the Twitter dataset, it fell short of EEGCN's performance on the Rest14 and Rest16 datasets. Specifically, EEGCN significantly surpasses other models in both Acc and F1 scores on the Rest14 dataset. In the Rest15 dataset, while RACL achieved the highest Acc score, its F1 score was relatively low at 59.85. For the Rest16 dataset, the Acc and F1 scores of TNet-LF were only marginally better than those of EEGCN. This suggests that the comprehensive consideration of both parent and child nodes is crucial, underscoring the importance of analyzing the entire graph structure rather than treating it as a directed graph. Additionally, the limited grammatical complexity of sentences in the Twitter dataset might have impacted the prediction results. Overall, the results demonstrate that EEGCN significantly enhances aspect-based sentiment analysis compared to other baseline models, particularly outperforming TNet-LF on the Rest16 dataset and surpassing ASGCN across all four benchmark datasets. This highlights the effectiveness of incorporating word dependencies to improve classification accuracy.

%-----------------------------------
\subsection{Effect of GCN Layers}
\label{sec:layers}

\begin{figure}[ht]  
\centering  
\includegraphics[width=0.65\hsize]{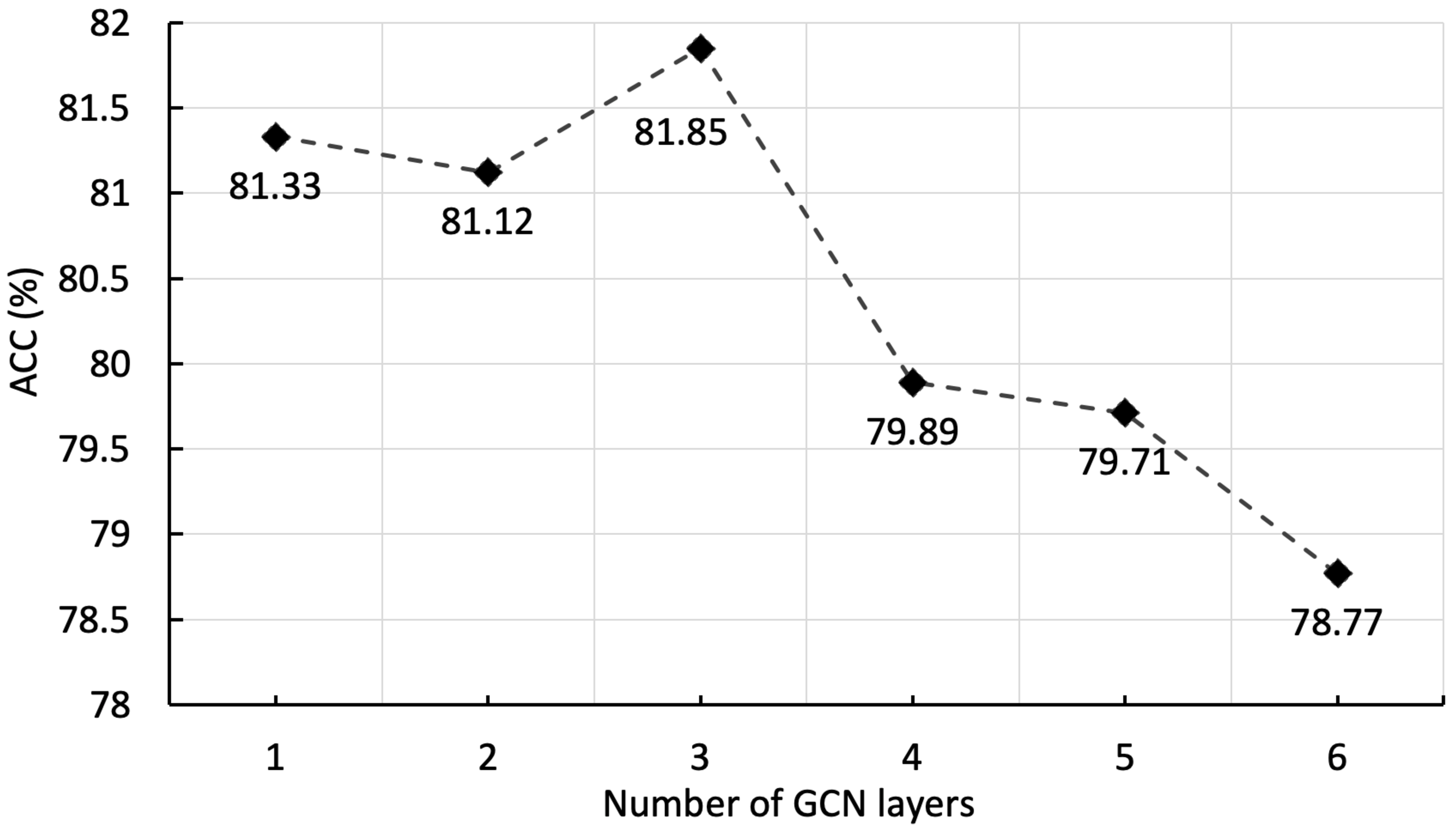}   
\caption{Change curve of Acc scores with the number of GCN layers.}   
\label{fig:acc} 
\end{figure}

\begin{figure}[ht]
\centering  
\includegraphics[width=0.655\hsize]{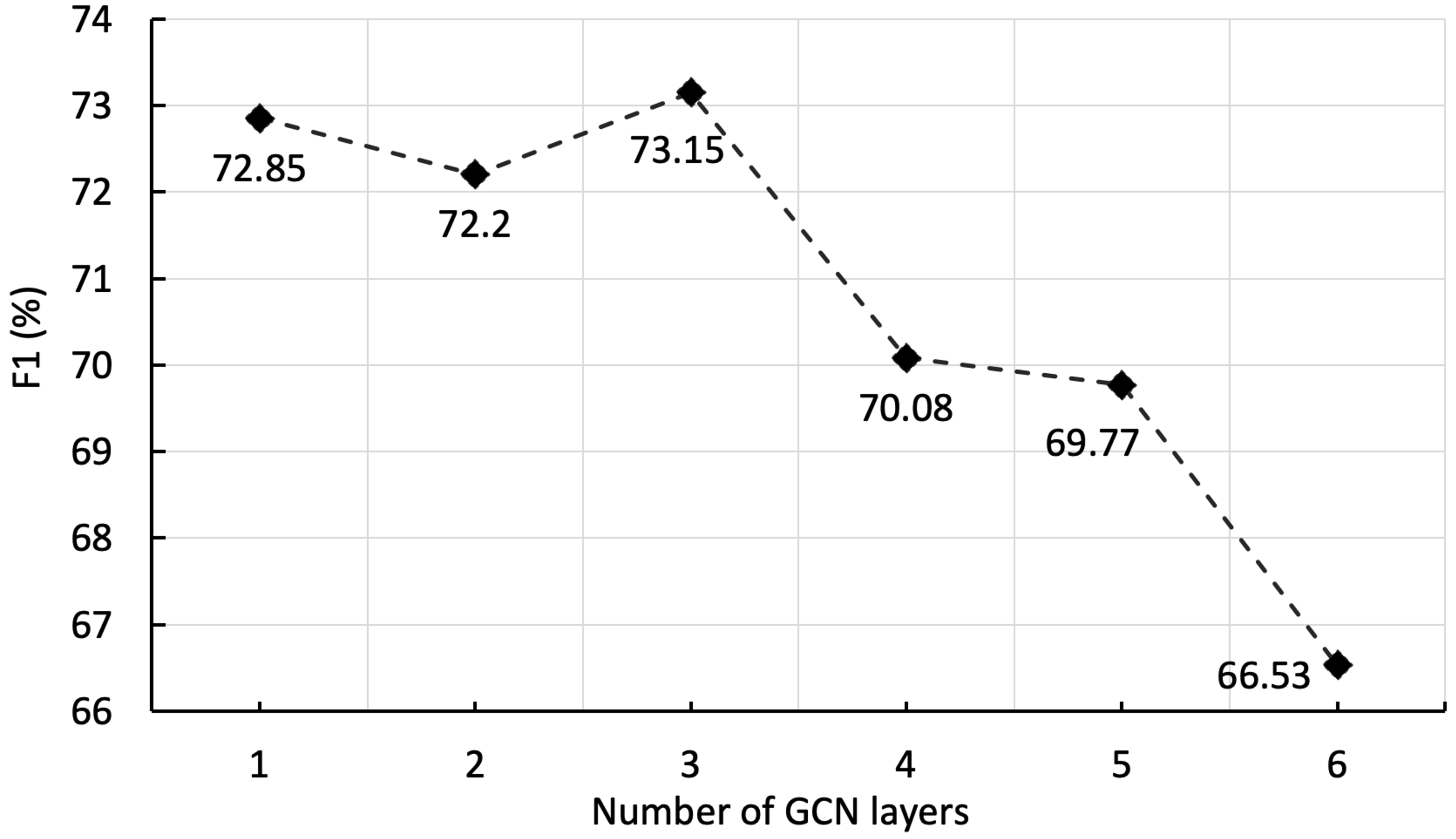}   
\caption{Change curve of F1 scores with the number of GCN layers.}   
\label{fig:f1} 
\end{figure}

Figures \ref{fig:acc} and \ref{fig:f1} illustrate how varying the number of GCN layers impacts the Accuracy (Acc) and Macro F1 (F1) scores of the EEGCN model on the Rest14 dataset. The X-axis represents the number of GCN layers, while the Y-axis shows the corresponding Acc and F1 scores. The results indicate that both the Acc and F1 scores initially increase with the addition of GCN layers, reaching peak values at 3 layers, with scores of 0.818 and 0.731, respectively. Beyond this point, further increases in the number of layers result in a decline in performance.

%-----------------------------------
\subsection{Ablation Studies}
\label{sec:ablation}
\begin{table*}[t]
\caption{Ablation experiments on syntactic dependencies ($\boldsymbol{D}$), edge weights ($\boldsymbol{EW}$), and bidirectional connectivity ($\boldsymbol{Bi}$) in the EEGCN model.} 
\centering
\begin{tabular}{@{}cccccc@{}}\toprule
Dataset & Measures & $\boldsymbol{D}$ w/o & $\boldsymbol{EW}$ w/o & $\boldsymbol{Bi}$ w/o & EEGCN \\\hline
\multirow{2}{*}{Twitter}&Acc (\%) $\uparrow$ & 71.53 & 71.39 & 71.80 & \colorbox[gray]{0.9}{72.40} \\
&F1 (\%) $\uparrow$ & 69.68 & 69.37 & 70.42 & \colorbox[gray]{0.9}{70.67} \\\hline
\multirow{2}{*}{Rest14}&Acc (\%) $\uparrow$ & 80.86 & 81.34 & 81.57 & \colorbox[gray]{0.9}{81.70} \\
&F1 (\%) $\uparrow$ & 72.19 & 71.70 & 73.35 & \colorbox[gray]{0.9}{73.63} \\\hline
\multirow{2}{*}{Rest15}&Acc (\%) $\uparrow$ & 79.34 & 79.15 & \colorbox[gray]{0.9}{80.44} & 79.34 \\
&F1 (\%) $\uparrow$ & 60.78 & 60.06 & 61.58 & \colorbox[gray]{0.9}{61.99} \\\hline
\multirow{2}{*}{Rest16}&Acc (\%) $\uparrow$ & 88.69 & 88.28 & 88.37 & \colorbox[gray]{0.9}{88.80} \\
&F1 (\%) $\uparrow$ & 66.64 & 67.03 & \colorbox[gray]{0.9}{69.20} & 68.96 \\\toprule
\end{tabular}
\label{tab:study}
\end{table*}

The proposed EEGCN model comprises three key components: the grammar dependency tree, the edge weight matrix, and the Bi-GCN. Ablation studies have been undertaken to assess the impact of each submodule on performance. Table \ref{tab:study} displays the Acc and F1 scores of the EEGCN model under varied conditions: without grammar dependency, with discrete edge weights, and minus the bidirectional GCN across four benchmark datasets. The findings highlight the significance of the grammar dependency technique, showcasing a notable enhancement in model performance. Models incorporating dependencies consistently outperformed those without dependencies across all datasets. Furthermore, utilizing the SDI edge weight matrix yielded superior results compared to traditional matrices with binary representations (0 and 1) on all datasets, underscoring the performance gains associated with SDI matrices. Moreover, the incorporation of the Bi-GCN in the EEGCN model demonstrated enhanced performance across six out of eight metrics compared to models without bidirectional components. Collectively, the grammar dependency tree, edge weight matrix enhancements, and Bi-GCN integration collectively contributed to the notable improvement in model performance.

%****************************************************************
\section{Conclusion}
\label{sec:conclusion}
This study outlined an enhancement to aspect-based sentiment analysis systems, known as EEGCN, by leveraging an edge-enhanced bidirectional GCN. The EEGCN model employed a Bi-LSTM to gain a profound understanding of text syntax and semantics. The hidden layer features obtained from the Bi-LSTM were used to construct syntactic dependency trees in the Bi-GCN. Additionally, EEGCN integrated a transformer encoder layer to extract global information from text, thereby improving its ability to capture nuanced features in lengthy textual data. The dependency parsing layer in the Bi-GCN facilitated learning connections between aspects and context based on the text's syntax. To refine the model further, an aspect-specific masking layer was implemented to reduce redundant information in the GCN's hidden layer, consequently boosting the accuracy of the aspect-based sentiment analysis system. Finally, EEGCN combined the outputs from the Bi-GCN and transformer layers, serving as inputs for the sentiment classification classifier.

Two primary limitations were identified in the EEGCN model. Firstly, the computation of the SDI, a significant contribution of the research, was conducted statistically. Future efforts aim to enhance the model by incorporating trainable vectors within the network to enrich the features associated with each edge. Secondly, the use of the transformer for feature enhancement raises questions about the most effective stage for integrating global information into the model. Future work intends to address these limitations by incorporating trainable vectors within the network to enable a more dynamic and context-specific computation of the SDI, thereby enriching edge features. Additionally, strategies will be explored to optimize the integration of global information from the transformer to enhance overall classification accuracy.

%****************************************************************
\section*{Acknowledgements}
This research was supported by the International Research Fellow of Japan Society for the Promotion of Science (Postdoctoral Fellowships for Research in Japan [Standard]).

%****************************************************************
\bibliography{refs}
\end{document}